\documentclass[letterpaper]{article} 
\usepackage{aaai24}  
\usepackage{times}  
\usepackage{helvet}  
\usepackage{courier}  
\usepackage[hyphens]{url}  
\usepackage{graphicx} 
\usepackage{natbib}  
\usepackage{caption} 
\usepackage{algorithm}
\usepackage{algorithmic}
\usepackage{amsthm, amsmath, amssymb}
\usepackage{multirow}
\usepackage{booktabs}
\usepackage{newfloat}
\usepackage{listings}
\DeclareCaptionStyle{ruled}{labelfont=normalfont,labelsep=colon,strut=off} 
\floatstyle{ruled}
\newfloat{listing}{tb}{lst}{}
\floatname{listing}{Listing}
\title{TF-CLIP: Learning Text-free CLIP for Video-based Person Re-Identification}
\author {
	Chenyang Yu\textsuperscript{\rm 1},
	Xuehu Liu\textsuperscript{\rm 2},
	Yingquan Wang\textsuperscript{\rm 1},
	Pingping Zhang\textsuperscript{\rm 3}\thanks{Corresponding author.},
	Huchuan Lu\textsuperscript{\rm 1,\rm 3, \rm 4}
	\\
}
\affiliations{
	\textsuperscript{\rm 1} School of Information and Communication Engineering, Dalian University of Technology, Dalian, China\\
	\textsuperscript{\rm 2} School of Computer Science and Artificial Intelligence, Wuhan University of Technology, Wuhan, China\\
	\textsuperscript{\rm 3} School of Future Technology, School of Artificial Intelligence, Dalian University of Technology, Dalian, China\\
	\textsuperscript{\rm 4} Ningbo Institute, Dalian University of Technology, Ningbo, China\\
	yuchenyang@mail.dlut.edu.cn;liuxuehu@whut.edu.cn;yingquan\_w95@mail.dlut.edu.cn;{\{zhpp, lhchuan\}@dlut.edu.cn}
}

\begin{document}

\maketitle

\begin{abstract}
Large-scale language-image pre-trained models (e.g., CLIP) have shown superior performances on many cross-modal retrieval tasks.
However, the problem of transferring the knowledge learned from such models to video-based person re-identification (ReID) has barely been explored.
In addition, there is a lack of decent text descriptions in current ReID benchmarks.
To address these issues, in this work, we propose a novel one-stage text-free CLIP-based learning framework named TF-CLIP for video-based person ReID.
More specifically, we extract the identity-specific sequence feature as the CLIP-Memory to replace the text feature.
Meanwhile, we design a Sequence-Specific Prompt (SSP) module to update the CLIP-Memory online.
To capture temporal information, we further propose a Temporal Memory Diffusion (TMD) module, which consists of two key components: Temporal Memory Construction (TMC) and Memory Diffusion (MD).
Technically, TMC allows the frame-level memories in a sequence to communicate with each other, and to extract temporal information based on the relations within the sequence.
MD further diffuses the temporal memories to each token in the original features to obtain more robust sequence features.
Extensive experiments demonstrate that our proposed method shows much better results than other state-of-the-art methods on MARS, LS-VID and iLIDS-VID.
\end{abstract}

\section{Introduction}
Video-based person Re-Identification (ReID)~\cite{liu2021watching,xu2017jointly,liu2015spatio} aims at re-identifying specific persons from videos across non-overlapping cameras.
Over the past decade, various approaches have been proposed to solve this challenging task, including CNN-based methods~\cite{zhang2020multi,fu2019sta,subramaniam2019co,chen2018video,dai2019video} and Transformer-based methods~\cite{liu2021video,tang2023multi,zang2022multidirection,wu2022cavit}.
However, existing works employ a unimodal framework, where the video encoders are trained to predict a fixed set of predefined labels.
Considering that the ReID task is in an open-set setting, such closed-set training setting limits the robustness of these methods.

Recently, the outstanding Contrastive Language-Image Pre-training (CLIP)~\cite{radford2021learning} has shown a great capability of learning robust representations.
Different from traditional unimodal frameworks, CLIP is trained on large-scale language-image pairs in a contrastive way.
Apparently, CLIP-based methods have achieved great success in video domains~\cite{ju2022prompting,xu2021videoclip,wang2021actionclip,ni2022expanding}.
However, the transfer and adaptation to video-based person ReID is not well explored.
A major challenge is that compared with other video tasks, video sequences in video-based person ReID only have integer index labels, and there is no text labels.
A straightforward solution is to annotate the natural language descriptions for existing video-based person ReID datasets.
Unfortunately, it is both time consuming and labour cost to annotate suitable text descriptions for fine-grained person videos.
\begin{figure}
	\centering
	\includegraphics[width=0.95\linewidth]{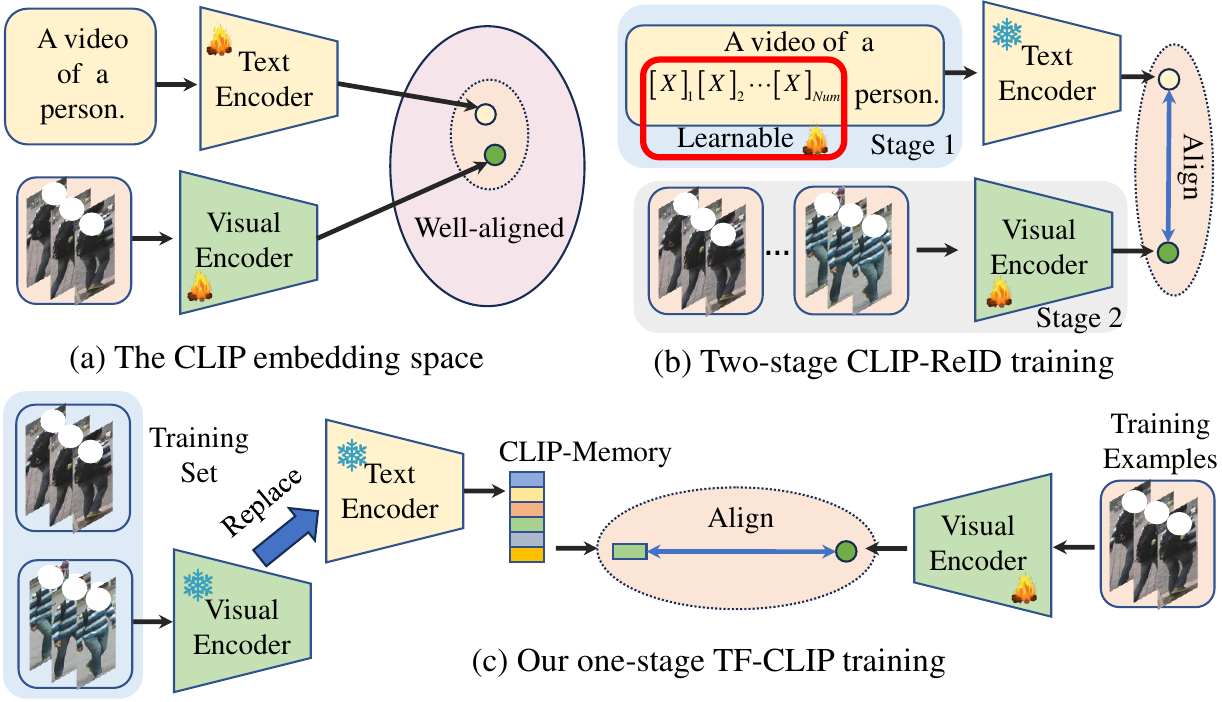}
	\caption{Comparison of CLIP-ReID and our one-stage text-free learning framework. (a) An illustration of the CLIP joint embedding space. (b) Two-stage CLIP-ReID training. (c) Our one-stage TF-CLIP training.
	}
	\label{fig:intro}
\end{figure}

To overcome these issues, inspired by recent prompt engineering researches, CLIP-ReID~\cite{li2022clip} is proposed for image-based person ReID.
As shown in Fig.~\ref{fig:intro} (b), CLIP-ReID fixes the text encoder and visual encoder in the first training stage, and optimizes a set of learnable text tokens to generate the label-specific text features.
Then, it uses the learned text features to update the visual encoder in the second training stage.
Although CLIP-ReID has achieved excellent performances, it does not consider the temporal information in video sequences.
For video-based person ReID, extracting robust temporal features is a critical step~\cite{gao2018revisiting}.
Thus, directly applying CLIP-ReID in video-based person ReID is sub-optimal.
On the other hand, the two-stage training strategy is not elegant enough and needs hyper-parameter tuning.

To address the aforementioned issues, we propose an one-stage text-free CLIP-based person ReID framework named TF-CLIP.
Specifically, as shown in Fig.~\ref{fig:intro} (c), we first use a pre-trained CLIP visual encoder for all video sequences of each identity to extract sequence features, and then take the average to obtain the identity-level feature denoted as CLIP-Memory.
In this form, we replace the text encoder and achieve text-free CLIP-based person ReID.
As shown in Fig.~\ref{fig:intro} (a), this is reasonable because the pre-trained CLIP text encoder and visual encoder are constrained by contrastive learning losses, where the outputs of the two encoders are aligned in a common feature space.
What's more, the essence of the first-stage training of CLIP-ReID is to align text prompts with the outputs of the pre-trained visual encoder.
Based on the above facts, we propose CLIP-Memory to address the lack of appropriate text descriptions when using pre-trained visual-language models.
Considering that the CLIP-Memory is fixed during the training process, the resulting method will be only optimized for a specific set of training identities, while ignoring the appearance diversity of the same identity.
Inspired by CoCoOp~\cite{zhou2022conditional}, we further design a Sequence-Specific Prompt (SSP) module, which updates this CLIP-Memory according to each video sequence when training.

On the other hand, we further propose a Temporal Memory Diffusion (TMD) module to capture temporal information.
TMD is composed of Temporal Memory Construction (TMC) and Memory Diffusion (MD).
Technically, it first takes the feature of each frame as input, and constructs a memory token.
Then, based on memory tokens, TMC is employed to capture temporal information within the sequence.
The updated memory token will store the context information of the video sequence.
Next, MD combines the obtained temporal memory tokens with the original features, and diffuses the temporal memory to each token.
Finally, we perform a Temporal Average Pooling (TAP) on the updated frame-level features to obtain robust sequence-level feature representations.
Extensive experiments on three public ReID benchmarks demonstrate that our approach achieves promising results over most of previous methods.

In summary, our contributions are three folds:
\begin{itemize}
	\item
	We propose a novel one-stage text-free CLIP-based learning framework named TF-CLIP for video-based person ReID.
	To our best knowledge, we are the first to extract identity-specific sequence features to replace the text features of CLIP.
	Meanwhile, we further design a Sequence-Specific Prompt (SSP) module to update the CLIP-Memory online.
	\item
	We propose a Temporal Memory Diffusion (TMD) module to capture temporal information.
	The frame-level memories in a sequence first communicate with each other to extract temporal information.
	The temporal information is then further diffused to each token, and finally aggregated to obtain more robust temporal features.
	\item
	Extensive experiments demonstrate that our proposed method shows superior performance over existing  methods on three video-based person ReID datasets, \emph{i.e.}, MARS, LS-VID and iLIDS-VID.
\end{itemize}
\begin{figure*}[t]
	\centering
	\includegraphics[width=0.9\textwidth]{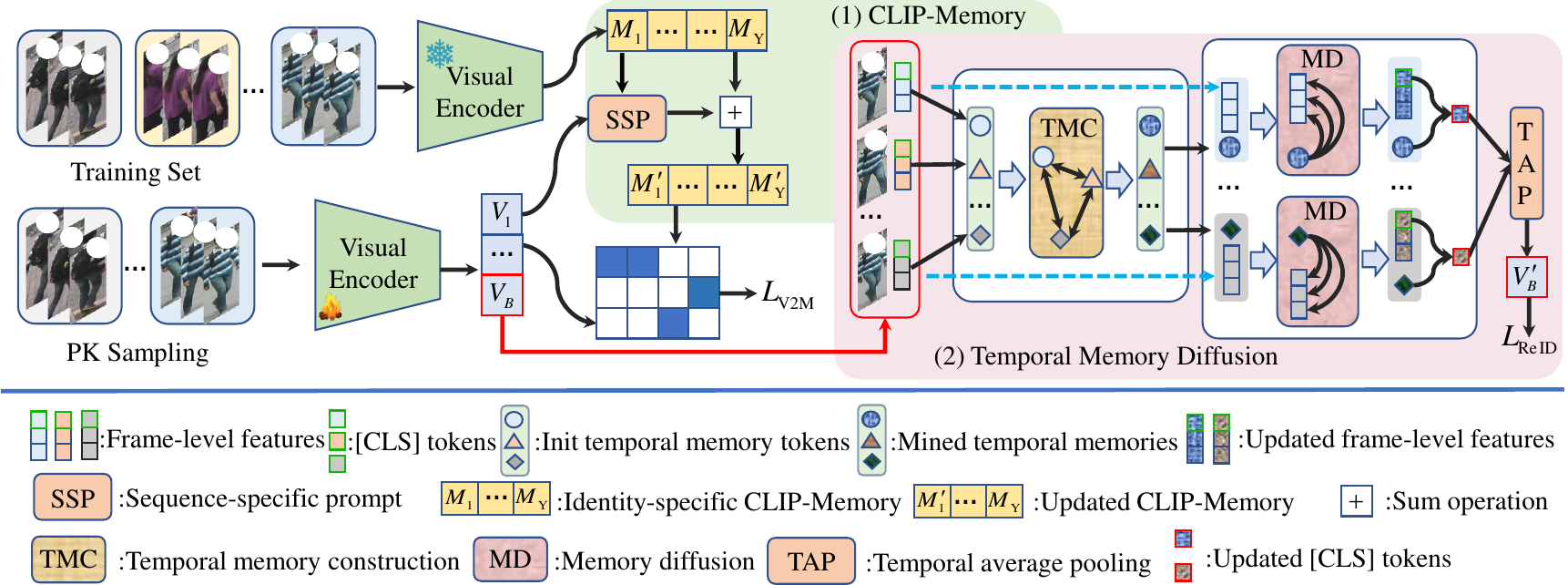}
	\caption{Illustration of the proposed TF-CLIP framework. It consists of two key modules, \emph{i.e.,} a CLIP-Memory module and a Temporal Memory Diffusion (TMD) module. (1) In the CLIP-Memory module, we first extract the identity-specific sequence feature as CLIP-Memory to replace the text feature.
	Meanwhile, a Sequence-Specific Prompt (SSP) module is proposed to update the CLIP-Memory online. (2) In TMD, Temporal Memory Construction (TMC) is first proposed to capture temporal information based on the relations within the sequence.
	Then, Memory Diffusion (MD) further diffuses the temporal information to each token in the original features to obtain more robust sequence features.
	}
	\label{fig:overall}
\end{figure*}
\section{Related Works}
\subsection{Video-based Person ReID}
Most of video-based person ReID methods exploit spatial and temporal cues for video representation learning.
Early studies use RNNs/LSTMs~\cite{mclaughlin2016recurrent,dai2018Video}, 3D convolutions~\cite{li2019multi,gu2020appearance}, temporal pooling~\cite{zheng2016mars,wu2018exploit} and attention mechanisms~\cite{liu2021watching,liu2023video}, to extract temporal features.
For example, Hou \emph{et al.}~\cite{hou2020temporal} propose a temporal complementary learning network to extract complementary features of consecutive video frames.
%
%
Different from these methods, we propose a Temporal Memory Diffusion (TMD) module to mine temporal memory and diffuse it to each frame.

In recent years, the vision Transformer has shown impressive results in person ReID compared with CNN-based methods.
Liu \emph{et al.}~\cite{liu2021video} propose a trigeminal Transformer in spatial, temporal and spatial-temporal views to obtain more cues.
He \emph{et al.}~\cite{he2021dense} propose a hybrid framework based on dense interaction learning, which utilizes both CNNs and attention mechanisms to enhance multi-grained spatio-temporal modeling.
%
%
%
Despite the impressive success of above approaches, they all employ a unimodal framework, where video encoders are trained to predict a fixed set of predefined labels.
In contrast, we propose a new paradigm based on a visual-language multimodal learning framework for video-based person ReID.
\subsection{Visual-language Learning}
Recently, visual-language joint learning methods~\cite{jia2021scaling,yuan2021florence} (\emph{e.g.}, CLIP) have demonstrated great potential in extracting generic visual representations.
For example, Rao \emph{et al.}~\cite{rao2022denseclip} propose DenseCLIP for dense prediction by implicitly and explicitly leveraging the pre-trained knowledge from CLIP.
Zhou \emph{et al.} propose CoOp~\cite{zhou2022learning} and CoCoOp~\cite{zhou2022conditional} for zero-shot image classification by prompt learning.
Khattak \emph{et al.}~\cite{khattak2022maple} propose multi-modal prompt learning for both vision and language branches to improve alignments between vision and language representations.
Li \emph{et al.}~\cite{li2022clip} utilize CLIP for image-based person ReID with a two-stage training strategy.
Different from the above methods, we design an one-stage training strategy to extend CLIP to the video-based person ReID task.
On the other hand, several studies also try to extend the existing CLIP model to the video domain.
Rasheed \emph{et al.}~\cite{rasheed2022fine} propose ViFi-CLIP to bridge the domain gap from images to videos.
Wang \emph{et al.}~\cite{wang2021actionclip} propose ActionCLIP for action recognition by designing appropriate prompts.
Ni \emph{et al.}~\cite{ni2022expanding} propose XCLIP and use a text prompt generation for better generalization.
Inspired by the successful applications of CLIP, we further design a Temporal Memory Diffusion (TMD) module for video-based person ReID.
\section{Brief Reviews of CLIP and CLIP-ReID}
CLIP consists of a visual encoder $\mathcal{V}(\cdot)$ and a text encoder $\mathcal{T}(\cdot)$, jointly trained with contrastive learning to respectively map the input image and text into a unified representation space.
Specifically, let $\{img_i,text_i \}_{i=1}^{B}$ be a set of $B$ training visual-language pairs within a batch, where $img_{i}$ is an image and $text_{i}$ is a corresponding text description.
CLIP uses the above two encoders and combines two linear projecting layers to encode images and texts into corresponding image features and text features.
Then, the similarity of the two features is calculated as:
\begin{equation}\label{s1}
	S(img_i,text_i)=\mathcal{J}_{v}(\mathcal{V}(img_i))\cdot \mathcal{J}_{t}(\mathcal{T}(text_i)),
\end{equation}
where $\mathcal{J}_{v}$ and $\mathcal{J}_{t}$ are linear layers to project features into a joint feature space.
Finally, two contrastive losses denoted as $L_{v2t}$ and $L_{t2v}$ are employed to train models~\cite{radford2021learning}.
Thanks to $L_{v2t}$ and $L_{t2v}$, the cross-modal features will be aligned.
In the application of downstream tasks, the text usually takes the form of prompt, such as “A photo/video of a \{class\}.” where “\{\}” is filled with a specific class name (\emph{e.g.}, cat).
Unfortunately, the labels in person ReID are integer indexes instead of specific texts.

To solve this problem, CLIP-ReID~\cite{li2022clip} introduces some identity-specific tokens to learn text descriptions.
Specifically, the text prompt becomes “A photo of a $[X]_1, [X]_2,\cdots,[X]_{Num}$ person”.
Then, in the additional training stage, CLIP-ReID fixes the parameters of $\mathcal{V}(\cdot)$ and $\mathcal{T}(\cdot)$, and uses the $L_{v2t}$ and $L_{t2v}$ to optimize the learnable identity-specific tokens.
In this way, CLIP-ReID learns a corresponding ambiguous text description for each identity.
Although CLIP-ReID achieves impressive results, it requires an additional training stage to learn text descriptions and does not consider the temporal information in videos.
%
\section{Our Method}
To deal with the aforementioned problems, we propose an one-stage text-free framework named TF-CLIP.
The proposed pipeline is depicted in Fig.~\ref{fig:overall}, which consists of two main modules, a CLIP-Memory module to replace the original text branch and a Temporal Memory Diffusion (TMD) module to extend CLIP to video-based person ReID.
\subsection{CLIP-Memory Module}
Once the computation of the visual branch is completed, the next step of previous CLIP-based methods~\cite{li2022clip,zhou2022learning,rao2022denseclip} is to generate the corresponding text features.
Unlike them, we argue that learning text descriptions is not necessary for CLIP-based person ReID method.
A CLIP-Memory can be used to replace the original text branch.
Specifically, given a video-based person ReID training set $\mathcal{D}=\{(x_{i}, y_{i})\}_{i=1}^{N_s}$ with labels $y_{i}\in \{1,\cdots,Y\}$, the total number of the person tracklets is $N_s$.
Taking identity $y_i$ as an example, we first employ the pre-trained CLIP visual encoder for all video sequences with the identity $y_i$ to extract identity-specific sequence features denoted as CLIP-Memory.

More specifically, given one of image sequences $\{I_t\}_{t=1}^T$ containing $T$ images belongings to the identity $y_i$, we split each frame $I_t\in \mathbb{R}^{H\times W\times 3}$ into $N_p$ non-overlapping patches $\{I_{t,i}\}_{i=1}^{N_p}\in \mathbb{R}^{P^2\times 3}$, in which $ H$ and $W$ represent the number of height and width, respectively.
$N_p=\frac{H}{P}\times \frac{W}{P}$ and $P$ is the patch size.
Then the patches are projected to token embeddings using a linear projection layer $E_{emb}\in \mathbb{R}^{3P^2\times D}$, where $D$ represents the dimension of each token.
Meanwhile, an extra learnable [CLS] token denoted as $CLS_t\in \mathbb{R}^{D}$ is added to the embedded tokens for each frame.
Thus, the input of the video encoder at the frame $t$ is given by:
\begin{equation}\label{embed1}
	z_t^{Patch}=[CLS_t, E_{emb}^{T}I_{t,1},\cdots,E_{emb}^{T}I_{t,N_p}] + e^{sp},
\end{equation}
where $e^{sp}$ is the spatial position embedding.
Then the above embeddings are sequentially processed through CLIP. At last, the frame-level representation is defined as,
\begin{equation}\label{embed2}
	z_t=\mathcal{V}^{*}(z_t^{Patch}),
\end{equation}
where the superscript $*$ indicates that the parameters of this visual encoder $\mathcal{V}$ are frozen during training.
What's more, the obtained class token $CLS_t$ is projected to a visual-language unified space via a Visual Project (VP) layer, $f_t=VP(z_{t,0})$,
where $f_t$ is the final representation of frame $t$.
In practice, $VP$ is implemented by a fully connected layer, represented as $VP=W_{VP}\in \mathbb{R}^{D\times d}$.
Finally, a Temporal Average Pooling (TAP) is employed to obtain the sequence-level feature $v_a$.
Once all the video sequence features belonging to the identity $y_i$ are obtained, the average of them can represent the identity-specific feature $M_{y_{i}}$ defined as CLIP-Memory:
\begin{equation}\label{clip-memory}
	M_{y_{i}}=\frac{1}{N_i}\sum_{v_a\in y_i}v_a,
\end{equation}
where $N_i$ is the number of sequences belonging to the identity $y_i$.
When traversing the entire training set, we can get an identity-specific CLIP-memory $M\in \mathbb{R}^{Y\times d}$.
Superior to previous text-associated CLIP-based methods, CLIP-Memory can exploit the image-text alignment potential of CLIP without using text information.
\begin{figure}[t]
	\centering
	\includegraphics[width=0.95\linewidth]{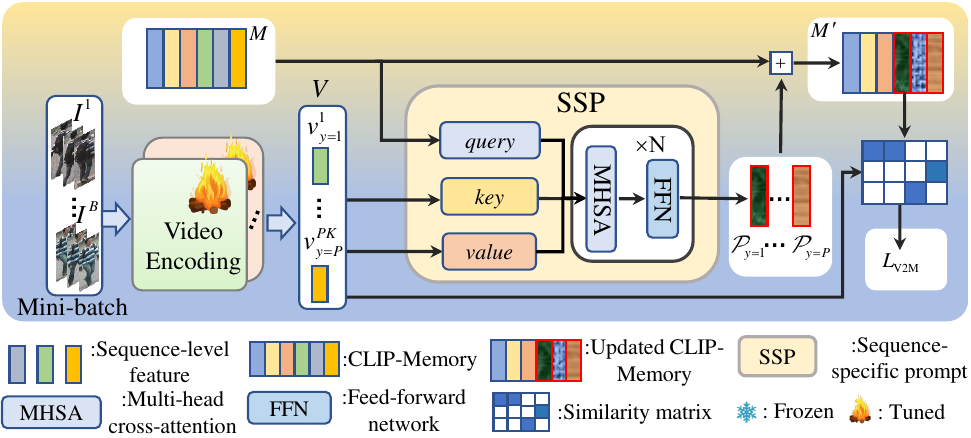}
	\caption{Illustration of our CLIP-Memory module.
	}
	\label{fig:clipmemory}
\end{figure}

\textbf{Sequence-Specific Prompt.}
In fact, if the CLIP-Memory is fixed during the training process, the framework will be only optimized for a specific set of training labels, while ignoring the appearance diversity of the same identity.
To address the above problem, as shown in Fig.~\ref{fig:clipmemory}, we introduce a novel module: Sequence-Specific Prompt (SSP).
The key idea is to generate a prompt conditioned on each input sequence to update the CLIP-Memory.

When training SSP, the PK sampling strategy~\cite{hermans2017defense} is employed to form a mini-batch $\{I^{b}\}_{b=1}^{B=P\times K}$, which contains $P$ different identities and $K$ video sequences for each identity.
After video encoding, the sequence-level features in the mini-batch can be expressed as $V=[v_{y=1}^{1},\cdots,v_{y=1}^{K},\cdots,v_{y=P}^{P\times K}]$.
Then, SSP takes the sequence-level feature $V$ and CLIP-Memory $M$ as inputs.
As shown in Fig.~\ref{fig:clipmemory}, the CLIP-Memory $M$ is used as $query$, and the sequence-level feature $V$ is used as $key$ and $value$.
Then, SSP employs the cross-attention mechanism to model the interactions between $query$, $key$ and $value$, which can be achieved by:
\begin{equation}\label{ssp}
	\mathcal{P}_{y=1,\cdots,y=P}=Transdecoder(query,key,value),
\end{equation}
where $\mathcal{P}$ represents the generated sequence-specific prompt.
There are $N$ blocks in $Transdecoder$, and each block consists of a Multi-Head Cross-Attention (MHCA) and a Feed-Forward Network (FFN).
Subscripts $y=1,\cdots,y=P$ represent the identities, which correspond to the index in the CLIP-Memory.
We then update the identity-specific CLIP-Memory through a residual connection:
\begin{equation}\label{updated}
	M_{y=1,\cdots,y=P}^{'}\leftarrow \mathcal{P}_{y=1,\cdots,y=P} + M_{y=1,\cdots,y=P}.
\end{equation}
In this way, CLIP-Memory can be updated online.
Finally, the updated CLIP-Memory $M^{'}\in \mathbb{R}^{Y\times d}$ and the sequence-level feature $V\in \mathbb{R}^{PK\times d}$ are employed for contrastive learning.
Compared with a fixed CLIP-Memory, SSP shifts the network's focus from a specific set of identities to each input sequence, thereby reducing overfitting and ultimately resulting in more discriminative features.
\subsection{Temporal Memory Diffusion}
\begin{figure*}
	\centering
	\includegraphics[width=0.9\textwidth]{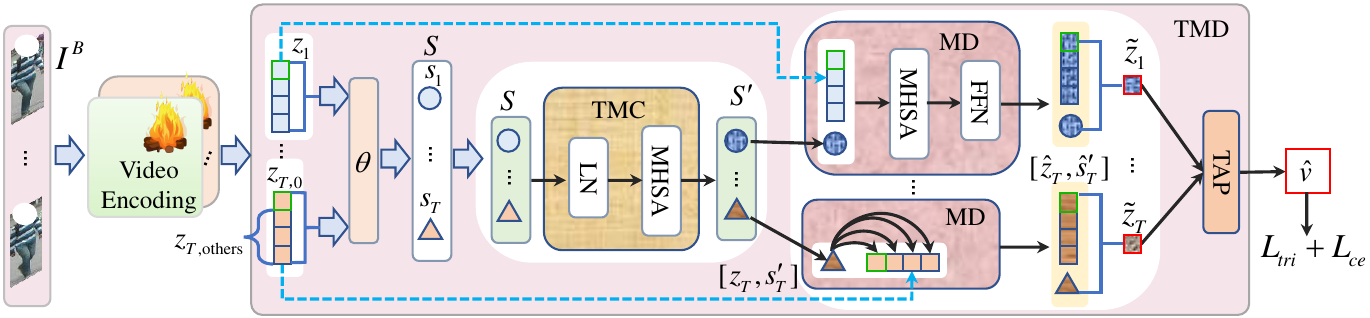}
	\caption{Illustration of the proposed Temporal Memory Diffusion (TMD) module.
	}
	\label{fig:TMD}
\end{figure*}

To capture temporal information, we further propose a Temporal Memory Diffusion (TMD) module.
As shown in Fig.~\ref{fig:TMD}, TMD is mainly composed of Temporal Memory Construction (TMC) and Memory Diffusion (MD).

\textbf{Temporal Memory Construction.}
Taking the last video sequence in the training batch as an example, which is defined as $I^{B}=\{I_1^{B},I_2^{B},\cdots,I_T^{B}\} $, where $B=P\times K$.
According to Eq.~\ref{embed1} and Eq.~\ref{embed2}, the representation at the frame level denoted as $z_t$ can be obtained through the video encoder.
During the standard training process of CLIP, the [CLS] token $z_{t,0}$ is used as the output of the visual encoder while the other tokens $z_{t, others}=[z_{t,1},\cdots, z_{t,N_p}]$ are usually neglected.
However, recent works~\cite{zeng2022not,chen2021crossvit,he2022transfg} have shown that $z_{t, others}$ still retains enough semantic information and can be used as a feature map.
Thus, Temporal Memory Construction (TMC) takes $z_t \in \mathbb{R}^{(N_p+1)\times D}$ as input.

As shown in Fig.~\ref{fig:TMD}, we first construct a memory token $s_t$ for each frame $t$ based on frame-level features, which is defined as follows:
\begin{equation}\label{temporal_memory}
	s_t=\theta(\frac{1}{N_p+1}\sum_{i=0}^{N_p}z_{t,i}),
\end{equation}
where we get the initial memory token based on the average of all tokens in each frame, and $\theta$ represents a liner projection.
Then, to capture the temporal cues in the sequence and transfer the information of each frame into the memory tokens, a Multi-Head Self-Attention (MHSA) is computed over the initial memory tokens $S=\{s_1,\cdots,s_T\}$:
\begin{equation}\label{temporal_memory+MHSA}
	S^{'}=MHSA(LN(S))+S,
\end{equation}
where $LN$ denotes Layer Normalization, and $S^{'}\in \mathbb{R}^{T\times D}$ are the temporal-aware memory tokens.
TMC allows the frame-level memories to communicate with each other, and extracts temporal information based on the relations obtained from neighbors.

\textbf{Memory Diffusion.} In order to make a better use of the temporal memory tokens, we design a diffusion-aggregation strategy named Memory Diffusion (MD).
As shown in Fig.~\ref{fig:TMD}, the temporal memory token $s_t^{'}\in \mathbb{R}^{1\times D}$ is first concatenated with each frame feature $z_t \in \mathbb{R}^{(N_p+1)\times D}$ to become $[z_t, s_t^{'}]\in \mathbb{R}^{(N_p+2)\times D}$,
where $[\cdot,\cdot]$ represents a concatenation of two features.
Then, a MHSA followed by a FFN layer is employed to diffuse the temporal information to each token in the frame, resulting in $[\hat{z}_t, \hat{s}_t^{'}]\in \mathbb{R}^{(N_p+2)\times D}$, which can be expressed as:
\begin{equation}\label{temporal_memory+MHSA+md}
	[\hat{z}_t, \hat{s}_t^{'}] = FFN(MHSA([z_t, s_t^{'}])) + [z_t, s_t^{'}].
\end{equation}
Finally, all tokens in each frame are aggregated to obtain frame-level features $\tilde{z}_t$ and a TAP is employed to further aggregate frame-level features into a sequence-level feature $\hat{v}$.
The above processes can be formulated as:
\begin{equation}\label{temporal_memory+MHSA+md+age}
	\tilde{z}_t=\frac{1}{N_p+2}(\sum_{i=0}^{N_p}\hat{z}_i+\hat{s}_t^{'}),
\end{equation}
\begin{equation}\label{temporal_memory+MHSA+md+age+TAP}
	\hat{v}=TAP([\tilde{z}_1,\cdots, \tilde{z}_T]).
\end{equation}
Superior to previous aggregation strategies, MD can diffuse the temporal memories to each token in the original features and obtain more robust sequence features.
\subsection{Optimization}
In this work, we utilize the video-to-memory contrastive loss denoted by $L_{V2M}$ to train the SSP module and the visual encoder in the CLIP-Memory.
It is worth noting that the visual encoder used to compute CLIP-Memory does not update during training.
Given the sequence-level feature $V=[v_1,\cdots,v_B]$ and the updated CLIP-Memory $M^{'}$,
it can be expressed as:
\begin{equation}\label{loss1}
	L_{V2M}(y_i)=-\frac{1}{\left| P(y_i)\right| }\sum_{p\in P(y_i)}log\frac{\exp(Cos(v_p,M_{y_i}^{'}))}{\sum_{j=1}^{B}\exp(Cos(v_j,M_{y_i}^{'}))}.
\end{equation}
where $P(y_i)$ represents all the positives for $M_{y_i}^{'}$ in the training batch and $\left| \cdot\right|$ is its cardinality.
$Cos$ represents the cosine similarity function.
$B$ represents the number of video sequences in a batch.

What's more, the triplet loss~\cite{hermans2017defense} denoted by $L_{tri}$ and the label smooth cross-entropy loss denoted by $L_{ce}$ are employed to optimize the visual encoder and the TMD module.
Finally, the losses used in our method are summarized as follows:
\begin{equation}\label{loss2}
	L_{total}=L_{V2M}+L_{tri}+L_{ce}.
\end{equation}
\section{Experiments}
\begin{table*}[]
	\centering
	\resizebox{0.92\textwidth}{!}{%
\begin{tabular}{ccccccccc}
	\hline
	\multirow{2}{*}{Methods}           & \multirow{2}{*}{Source}      & \multirow{2}{*}{Backbone} & \multicolumn{2}{c}{MARS}            & \multicolumn{2}{c}{LS-VID}          & \multicolumn{2}{c}{iLIDS-VID}       \\
	&                              &                           & mAP              & Rank-1           & mAP              & Rank-1           & Rank-1           & Rank-5           \\ \hline
	STMP~\cite{liu2019spatial}         & AAAI19                       & C                         & 72.7             & 84.4             & 39.1             & 56.8             & 84.3             & 96.8             \\
	M3D~\cite{li2019multi}             & AAAI19                       & C                         & 74.1             & 84.4             & 40.1             & 57.7             & 74.0             & 94.3             \\
	GLTR~\cite{li2019global}           & ICCV19                       & C                         & 78.5             & 87.0             & 44.3             & 63.1             & 86.0             & 98.0             \\
	TCLNet~\cite{hou2020temporal}      & ECCV20                       & C                         & 85.1             & 89.8             & 70.3             & 81.5             & 86.6             & -                \\
	MGH~\cite{yan2020learning}         & CVPR20                       & C                         & 85.8             & 90.0             & 61.8             & 79.6             & 85.6             & 97.1             \\
	GRL~\cite{liu2021watching}         & CVPR21                       & C                         & 84.8             & 91.0             & -                & -                & 90.4             & 98.3             \\
	BiCnet-TKS~\cite{hou2021bicnet}    & CVPR21                       & C                         & 86.0             & 90.2             & 75.1             & 84.6             & -                & -                \\
	CTL~\cite{liu2021spatial}          & CVPR21                       & C                         & 86.7             & 91.4             & -                & -                & 89.7             & 97.0             \\
	STMN~\cite{eom2021video}           & ICCV21                       & C                         & 84.5             & 90.5             & 69.2             & 82.1             & -                & -                \\
	PSTA~\cite{wang2021pyramid}        & ICCV21                       & C                         & 85.8             & 91.5             & -                & -                & 91.5             & 98.1             \\
	DIL~\cite{he2021dense}             & ICCV21                       & T                         & 87.0             & 90.8             & -                & -                & 92.0             & 98.0             \\
	STT~\cite{zhang2021spatiotemporal} & Arxiv21                      & T                         & 86.3             & 88.7             & 78.0             & 87.5             & 87.5             & 95.0             \\
	TMT~\cite{liu2021video}            & Arxiv21                      & T                         & 85.8             & 91.2             & -                & -                & 91.3             & 98.6             \\
	CAVIT~\cite{wu2022cavit}           & ECCV22                       & T                         & 87.2             & 90.8             & 79.2             & 89.2 & \underline{93.3} & 98.0             \\
	SINet~\cite{bai2022salient}        & CVPR22                       & C                         & 86.2             & 91.0             & 79.6 & 87.4             & 92.5             & -                \\
	MFA~\cite{gu2022motion}            & TIP22                        & C                         & 85.0             & 90.4             & 78.9             & 88.2             & \underline{93.3} & \underline{98.7} \\
	DCCT~\cite{liu2023deeply}          & TNNLS23                      & T                         & \underline{87.5} & \underline{92.3} & -                & -                & 91.7             & 98.6             \\
	LSTRL~\cite{liu2023video}  & ICIG23 & C       & 86.8             & 91.6             & \underline{82.4}             & \underline{89.8}             & 92.2             & 98.6             \\
	\textbf{TF-CLIP(Ours)}             &                              & T                         & \textbf{89.4}    & \textbf{93.0}    & \textbf{83.8}    & \textbf{90.4}    & \textbf{94.5}    & \textbf{99.1}    \\ \hline
\end{tabular}
	}
\caption{Comparison with typical CNN-based (C) and Transformer-based (T) methods on MARS, LS-VID and iLIDS-VID.}
\label{tab:Sota1}
\end{table*}
\begin{table*}[t]
	\begin{center}
		\resizebox{0.92\textwidth}{!}
		{
\begin{tabular}{cccccccccccc}
	\hline
	\multirow{2}{*}{Model} & \multicolumn{3}{c}{Components} & \multirow{2}{*}{Params(M)} & \multirow{2}{*}{FLOPs(G)}                    & \multicolumn{3}{c}{MARS} & \multicolumn{3}{c}{LS-VID} \\
	& CLIP-M     & SSP        & TMD    &                  &        & mAP   & Rank-1  & Rank-5 & mAP    & Rank-1  & Rank-5  \\ \hline
	1                      & $\times$   & $\times$   & $\times$   & 126.78    & 14.24    & 88.1  & 91.7    & 97.4   & 80.6   & 88.8    & 96.3    \\
	2                      & \checkmark & $\times$   & $\times$   & 86.94     & 11.26    & 88.4  & 91.6    & 97.9   & 80.3   & 87.7    & 95.7    \\
	3                      & \checkmark & \checkmark & $\times$   & 94.21     & 12.11    & 88.8  & 92.1    & 97.6   & 81.3   & 89.9    & 96.2    \\
	4                      & \checkmark & $\times$   & \checkmark & 97.02     & 11.72    & 89.2  & 92.3    & 97.7   & 81.6   & 88.9    & 96.4    \\
	5                      & $\times$          & $\times$        & \checkmark & 136.86    & 14.70    & 89.3  & 92.7    & 97.8   & 82.3   & 90.3    & 96.7    \\
	6                      & \checkmark & \checkmark & \checkmark & 104.26    & 12.53    & 89.4  & 93.0    & 97.9   & 83.8   & 90.4    & 97.1    \\ \hline
\end{tabular}
		}
	\end{center}
	\caption{Comparison of different components and the computational cost on MARS and LS-VID.}
	\label{table:abla}
\end{table*}
\subsection{Datasets and Evaluation Protocols}
We evaluate our proposed approach on three video-based person ReID benchmarks, including MARS~\cite{zheng2016mars}, LS-VID~\cite{li2019global} and iLIDS-VID~\cite{wang2014person}.
MARS is a large-scale benchmark which contains 20,715 videos with 1,261 identities.
LS-VID is a new benchmark which collects 3,772 identities and includes 14,943 video tracklets captured by 15 cameras.
iLIDS-VID is a small-scale dataset with 600 videos of 300 identities.
In addition, we follow common practices and adopt the Cumulative Matching Characteristic (CMC) at Rank-k and mean Average Precision (mAP) to measure the performance.
\subsection{Experiment Settings}
Our model is implemented on the PyTorch platform and trained with one NVIDIA Tesla A30 GPU (24G memory).
The ViT-B/16 from CLIP~\cite{radford2021learning} is used as the visual encoder's backbone.
The number of blocks in SPP is set to $N$=2.
During training, we sample 8 frames from each video sequence and each frame is resized to 256$\times$128.
In each mini-batch, we sample 4 identities, each with 4 tracklets.
Thus, the number of images in a batch is $4\times4\times8$=128.
We also adopt random flipping and random erasing~\cite{zhong2020random} for data augmentation.
We train our framework for 60 epochs in total by the Adam optimizer~\cite{kingma2014adam}.
Following CLIP-ReID~\cite{li2022clip}, we first warm up the model for 10 epochs with a linearly growing learning rate from 5 $\times 10^{-7}$ to 5 $\times 10^{-6}$.
Then, the learning rate is divided by 10 at the 30th and 50th epochs.
The original sequence-level feature $v\in \mathbb{R}^{1\times 512}$ and the aggregated feature $\hat{v}\in \mathbb{R}^{1\times 768}$ are concatenated to obtain the final video representation during testing.
The Euclidean distance is employed as the distance metric for ranking.
The code is available at https://github.com/AsuradaYuci/TF-CLIP.
\subsection{Comparison with State-of-the-arts}
In this section, we compare our methods with other methods on three video-based person ReID benchmarks.
The results are shown in Tab.~\ref{tab:Sota1}.
These experimental results clearly confirm the superiority and effectiveness of the proposed method on large-scale ReID datasets.

\textbf{Results on MARS}.
Tab.~\ref{tab:Sota1} shows that our proposed method achieves the best mAP of 89.4\% and the best Rank-1 of 93.0\% on MARS.
We can observe that our method shows much better results than other methods.
We note that most recent works (\emph{e.g.}, DIL~\cite{he2021dense}, CAVIT~\cite{wu2022cavit}, etc.) are developed with Transformers.
However, they are all based on unimodal pre-training.
Different from them, we propose TF-CLIP based on multi-modal pre-training.
Thus, our method surpasses CAVIT by 2.2\% and 2.2\% in terms of mAP and Rank-1 accuracy on MARS.
\begin{table*}[!t]
	\centering
	\begin{minipage}{0.34\textwidth}
		\centering
		\begin{tabular}{cccc}
			\toprule
			\multirow{2}{*}{Methods} & \multicolumn{2}{c}{MARS} & Params(M)/ \\ \cmidrule(r){2-3}
			& mAP        & Rank-1      & FLOPs(G)  \\ \midrule
			TAP                      & 88.4       & 91.6       & 86.94/11.26     \\
			Conv1D                    & 88.1       & 91.8       & 87.31/11.26   \\
			Transf$_{cls}$           & 87.5       & 91.5        & 106.23/11.41    \\
			Transf                   & 88.2       & 91.4       & 106.22/11.39    \\
			TMD                      & 89.2       & 92.3        & 97.02/11.72   \\ \bottomrule
		\end{tabular}
		\caption{Comparison of different temporal fusion methods.}
		\label{tab:abla3}
	\end{minipage}\quad
	\begin{minipage}{0.345\textwidth}
		\centering
		\begin{tabular}{cccc}
			\toprule
			\multirow{2}{*}{$N$} & \multicolumn{2}{c}{MARS} & Params(M)/  \\ \cmidrule(r){2-3}
			& mAP        & Rank-1     & FLOPs(G)  \\ \midrule
			1 & 88.5       & 91.9     & 91.06/11.68      \\
			2 & 88.8       & 92.1     & 94.21/12.11      \\
			3 & 87.4       & 92.1     & 97.36/12.53      \\
			4 & 88.4       & 92.3     & 100.51/12.95     \\ \bottomrule
		\end{tabular}
		\caption{Effect of different layers in SSP.}
		\label{tab:abla4}
	\end{minipage}\quad
	\begin{minipage}{0.275\textwidth}
		\centering
		
		\begin{tabular}{ccc}
			\toprule
			\multirow{2}{*}{Methods} & \multicolumn{2}{c}{MARS} \\ \cmidrule(r){2-3}
			& mAP        & Rank-1      \\ \midrule
			$Model2$  & 88.4       & 91.6        \\
			-wo CLIP-M & 84.9       & 90.2        \\ \hline
			$Model4$  & 89.2       & 92.3        \\
			-wo CLIP-M & 87.3       & 91.0        \\ \bottomrule
		\end{tabular}
		\caption{Comparison w/wo CLIP-Memory.}
		\label{tab:abla5}
	\end{minipage}
\end{table*}

\textbf{Results on LS-VID}.
It is observed that our method outperforms other state-of-the-arts on LS-VID.
Among them, STMN~\cite{eom2021video} also uses a temporal memory, which attains 82.1\% Rank-1 accuracy.
Instead, we propose the TMD module to generate the temporal memory in the sequence.
As a result, our method achieves 90.4\% Rank-1 accuracy, which surpasses STMN by 8.3\%.

\textbf{Results on iLIDS-VID}.
As can be observed, Tab.~\ref{tab:Sota1} also demonstrates the significant superiority of the proposed model over existing methods on iLIDS-VID.
Specially, our method achieves the best Rank-1 and Rank-5 accuracy of 94.2\% and 99.1\%, respectively.
It surpasses the previous best approaches MFA~\cite{gu2022motion} and CAVIT~\cite{wu2022cavit} by 1.2\% in terms of Rank-1 accuracy.
\subsection{Ablation Study}
To verify the impact of each component in our model, we conduct several experiments on MARS and LS-VID, and show compared results in Tab.~\ref{table:abla}.
``CLIP-M'' stands for CLIP-Memory.
$Model1$ means that CLIP-ReID is directly applied to video-based person ReID as the baseline, in which a TAP is employed to obtain sequence-level features.

\textbf{Effectiveness of CLIP-Memory.}
As can be seen from the first two rows in Tab.~\ref{table:abla}, $Model2$ obtains a higher Rank-5 and mAP (by 0.5\% and 0.3\%) than $Model1$ on MARS.
Meanwhile, competitive results are achieved on LS-VID.
The above results clearly demonstrate that it is feasible to use the CLIP-Memory to replace the text branch, which can inspire the subsequent extension of CLIP to other downstream tasks where text information is missing.

\textbf{Effectiveness of SSP.} %
As shown in Tab.~\ref{table:abla}, compared with $Model2$, $Model3$ with SPP brings 0.5\% mAP and 0.4\% Rank-1 accuracy gains on MARS, respectively.
What's more, compared with $Model4$,  $Model6$ also improves the Rank-1 accuracy by 1.5\% on LS-VID.
These results clearly demonstrate that our proposed SSP can indeed improve the performance.
A reasonable explanation for this improvement is that updating the CLIP-Memory online helps the network to learn a more discriminative representation.

\textbf{Effectiveness of TMD.}
As shown in Tab.~\ref{table:abla}, the proposed TMD improves the performance remarkably.
Compared with $Model2$, $Model4$ using TMD brings 0.8\% mAP and 0.7\% Rank-1 accuracy gains on MARS, respectively.
What's more, compared with $Model3$, $Model6$ also brings 0.6\% and 2.5\% Rank-1 accuracy gains on MARS and LS-VID, respectively.
The main reason is that our proposed TMD can capture temporal information in the sequence and obtain more robust sequence-level features.

\textbf{Comparison of different temporal fusion methods.}  %
Following ActionCLIP~\cite{wang2021actionclip}, we further investigate the impact of different temporal fusion methods with $Model2$ on MARS.
The experimental results are shown in Tab.~\ref{tab:abla3}.
It can be observed that the proposed TMD is more suitable for video-based person ReID.
For example, our method achieves 89.2\% Rank-1 accuracy on MARS, which surpasses one-dimensional convolution (Conv1D) by 1.1\%.
The reason is that our proposed TMD not only extracts temporal information in the sequence, but also passes it to each token in the frame.

\textbf{The effect of different layers in SSP.}
As shown in Tab.~\ref{tab:abla4}, we also carry out experiments to investigate the effect of different layers in SSP with $Model3$ on MARS.
We can see that our method is not sensitive to this hyper-parameter.
To balance the computation and performance, we finally choose $N=2$, which achieves the best mAP accuracy of 88.8\%.

\textbf{The necessity of CLIP-Memory.}
In fact, we can train the proposed framework without using CLIP-Memory.
The experimental results based on $Model2$ and $Model4$ are shown in Tab.~\ref{tab:abla5}.
It can be observed that not using CLIP-Memory will lead to a significant decrease in performance.
For example, the unimodal baseline method achieves only 84.9\% mAP accuracy on MARS, which is 3.5\% lower than $Model2$.
A plausible explanation for this performance drop is that the person ReID dataset is too small to adequately fine-tune the CLIP visual encoder when trained with unimodality.
This reflects the superiority of multimodal training and the necessity of our CLIP-Memory.
\subsection{Visualization}
\begin{figure}[t]
	\centering
	\includegraphics[width=0.98\columnwidth]{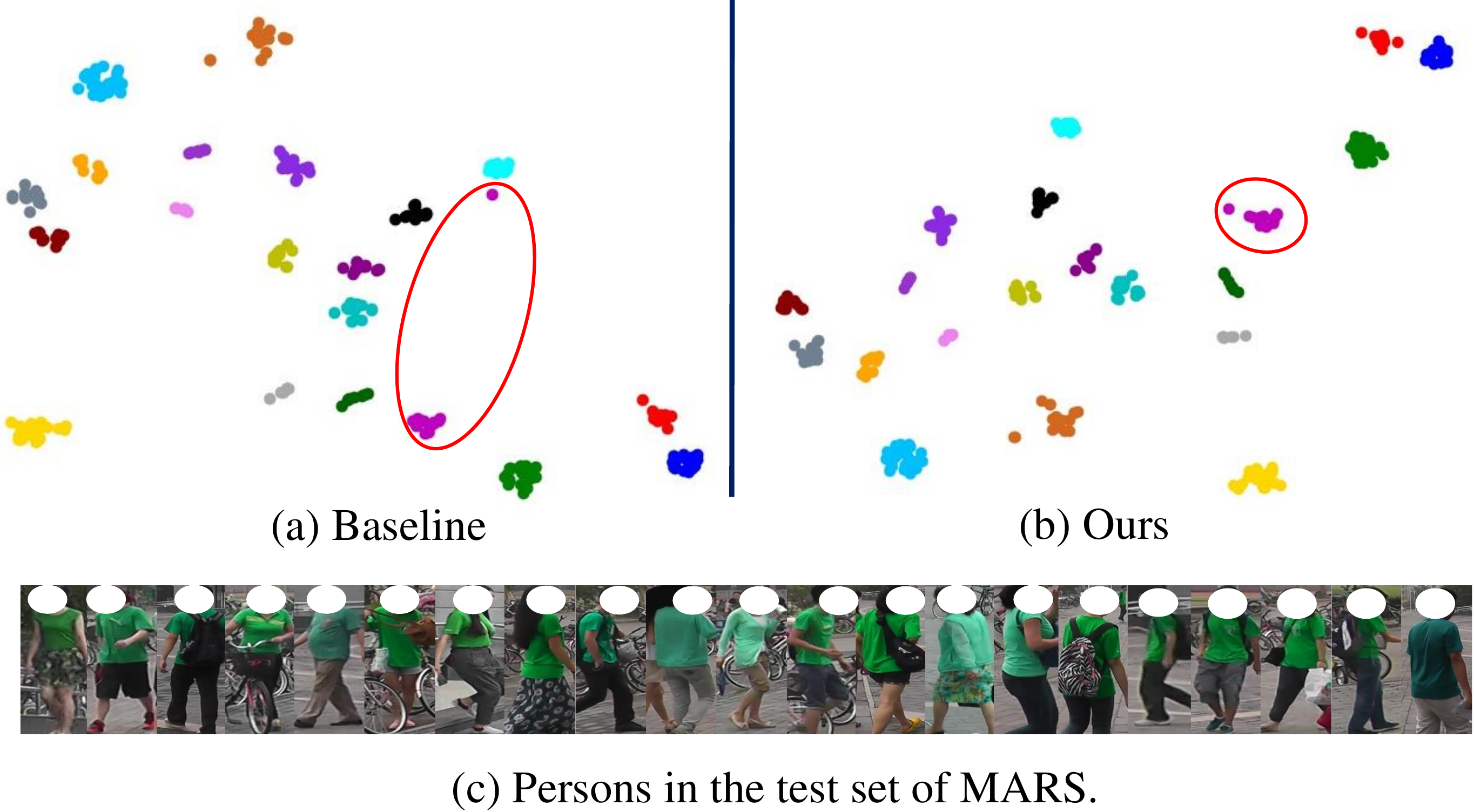}
	\caption{t-SNE visualization of the baseline and our method on the MARS test set. Different colored dots represent different identities. Best viewed in color.}
	\label{fig:tsne}
\end{figure}

To better understand the effect of the proposed TF-CLIP, some persons with similar appearances from MARS are chosen to visualize the feature distribution by t-SNE~\cite{van2008visualizing}.
As shown in Fig.~\ref{fig:tsne} (c), these persons are wearing green clothes with small inter-person variations.
Comparing Fig.~\ref{fig:tsne} (a) and (b), we can find that the proposed TF-CLIP indeed helps to learn more discriminative embeddings, with which the intra-person variance is minimized and the intra-person variance is maximized.
\section{Conclusion}
In this paper, we explore the application of vision-language pre-trained model to video-based person ReID without suitable textual descriptions.
Specifically, we propose a novel one-stage text-free CLIP-based framework named TF-CLIP.
To achieve text-free purpose, we propose CLIP-Memory to extract identity-specific sequence features to replace the text features.
Meanwhile, we design a Sequence-Specific Prompt (SSP) to update the CLIP-Memory online.
To capture temporal information, we further propose a Temporal Memory Diffusion (TMD) module.
It first constructs frame-level memories and lets them communicate with each other in the sequence to extract the temporal information.
Then the temporal memory is diffused into each token of the original frame, and aggregated to obtain more robust sequence features.
Extensive experiments on three public ReID datasets demonstrate the effectiveness and superiority of our method.

\section{Acknowledgments}
This work was supported in part by the National Key Research and Development Program of China (No. 2018AAA0102001), National Natural Science Foundation of China (No. 62101092) and Fundamental Research Funds for the Central Universities (No. DUT22QN228).

\bibliography{aaai24}

\end{document}